\documentclass[10pt,twocolumn,letterpaper]{article}

\usepackage{iccv}
\usepackage{times}
\usepackage{epsfig}
\usepackage{graphicx}
\usepackage{amsmath}
\usepackage{amssymb}
\usepackage{bm}
\usepackage{bbm}
\usepackage{multirow}
\usepackage{bbding}
\usepackage{subfigure}
\usepackage{xcolor}

\newcommand{\lyx}[1]{\textcolor{black}{#1}}
\newcommand{\js}[1]{\textcolor{black}{#1}}

\usepackage[pagebackref=true,breaklinks=true,colorlinks,bookmarks=false]{hyperref}

\iccvfinalcopy 

\def\iccvPaperID{1288} 

\ificcvfinal\fi

\begin{document}

\title{Enhancing Self-supervised Video Representation Learning via \\Multi-level Feature Optimization}

\author{Rui Qian\textsuperscript{1}, Yuxi Li\textsuperscript{1,2}, Huabin Liu\textsuperscript{1}, John See\textsuperscript{3}, Shuangrui Ding\textsuperscript{1}, Xian Liu\textsuperscript{4}, Dian Li\textsuperscript{5}, Weiyao Lin\textsuperscript{1}\thanks{Corresponding author. Email: wylin@sjtu.edu.cn}\\
\textsuperscript{1}Shanghai Jiao Tong Univeristy, \textsuperscript{2}Tencent Youtu Lab, \textsuperscript{3}Heriot-Watt University\\ \textsuperscript{4}Zhejiang University, \textsuperscript{5}Tencent PCG\\
\tt\{qrui9911,huabinliu,dsr1212,wylin\}@sjtu.edu.cn\\
\tt\{yukiyxli,goodli\}@tencent.com,j.see@hw.ac.uk,alvinliu@zju.edu.cn}

\maketitle
\ificcvfinal\thispagestyle{empty}\fi

\begin{abstract}
\js{The crux of} self-supervised video representation learning is to build \lyx{general features} from unlabeled videos. \js{However,} most \js{recent} works \js{have mainly} \lyx{focused on} high-level semantics and neglected lower-level \lyx{representations and their temporal relationship which} are crucial for \lyx{general video understanding}. To address these challenges, this paper proposes a multi-level feature optimization framework to improve the \lyx{generalization} and temporal modeling ability of learned video representations. Concretely, \lyx{high-level features obtained from} naive and prototypical contrastive learning \lyx{are utilized to build} distribution graphs, \lyx{guiding the process of} low-level and mid-level feature learning. \js{We also devise} 
a simple temporal modeling module from multi-level features 
to enhance motion pattern learning. Experiments demonstrate that multi-level feature optimization with the graph constraint \lyx{and temporal modeling} \js{can greatly} improve the \lyx{representation ability} in video \lyx{understanding}. Code is available \href{https://github.com/shvdiwnkozbw/Video-Representation-via-Multi-level-Optimization}{here}.
\end{abstract}

\section{Introduction}

Video representation learning has been a fundamental problem in computer vision to solve a series of video analysis tasks, \emph{e.g.}, action recognition and detection~\cite{carreira2017quo,xie2018rethinking,caba2015activitynet,gu2018ava,lin2020human,zhang2021regional}, video retrieval~\cite{liu2019use,miech2019howto100m}, video caption~\cite{wang2018reconstruction,pan2017video}, and etc. To address this problem, some large-scale human annotated datasets, \emph{e.g.}, Kinetics~\cite{carreira2017quo}, ActivityNet~\cite{caba2015activitynet}, YouTube-8M~\cite{abu2016youtube}, are developed to facilitate video understanding in specific downstream tasks. However, human labeling on videos is expensive, and fully-supervised methods fail to leverage massive unlabeled video data. Therefore, it is significant to develop unsupervised video representation learning without resorting to \lyx{manual labeling}.

\begin{figure}
    \centering
    \subfigure[]{
    \includegraphics[width=0.22\linewidth]{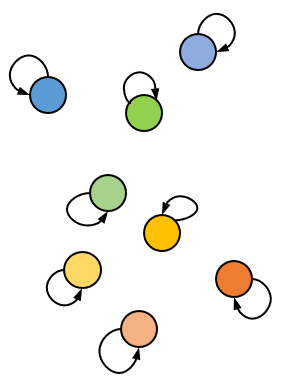}
    \label{a}}
    \subfigure[]{
    \includegraphics[width=0.22\linewidth]{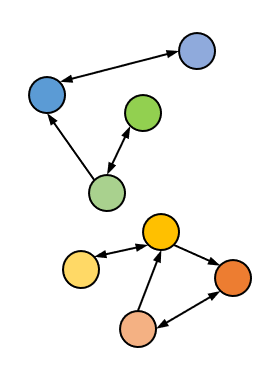}
    \label{b}}
    \subfigure[]{
    \includegraphics[width=0.22\linewidth]{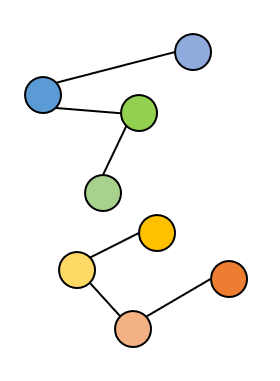}
    \label{c}}
    \subfigure[]{
    \includegraphics[width=0.22\linewidth]{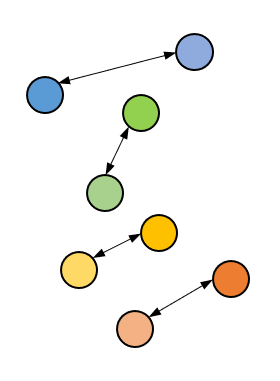}
    \label{d}}
    \caption{Graph presentation of four conditions. The nodes denote different samples, the edges present sample-wise relationships, and different colors indicate different characteristics, \emph{e.g.}, appearance, motion and semantic. Fig.~\ref{a} one hot label in InfoNCE loss, we use self-loop to present only the instance with its augmented view are regarded as positive. Fig.~\ref{b} instance-wise similarity distribution, measured by cosine similarity in embedding space\protect\footnotemark[1]. We use arrows to show the samples with similarity above threshold. Fig.~\ref{c} semantic-wise distribution, we connect samples of the same category. Fig.~\ref{d} comprehensive distribution formulated by the intersection of the former two. Note that we omit self-loops in the last three for concise presentation.}
    \label{fig:teaser}
    \vspace{-1.0em}
\end{figure}
\footnotetext[1]{The instance-wise similarity distribution is asymmetric due to the normalization, it is a directed graph.}

To achieve this goal, early works designed various pretext tasks to uncover effective supervision from video sequences ~\cite{benaim2020speednet,misra2016shuffle,kim2019self,jenni2020video,xu2019self,wang2020statistic}. 
Recently, contrastive learning has shown to be powerful in image representation learning~\cite{hjelm2018learning,oord2018representation,tian2019contrastive,chen2020simple,he2020momentum,ye2019unsupervised}. It encourages augmentation invariant representations by leveraging instance discrimination to attract augmented samples of the same instance and repel those of different instances. Later, beyond naive instance discrimination, inter-image relationships \lyx{and semantic structures are proved helpful for learning high-quality} representations~\cite{li2020prototypical,wang2020unsupervised}. To expand this pipeline to video domain, diverse spatiotemporal augmentation techniques are proposed to construct contrastive pairs and enhance motion modeling ~\cite{gordon2020watching,qian2020spatiotemporal,wang2020self,yao2020seco,dong2020motion}. Some works used contrastive learning to form temporal cycle or make future prediction to boost dense spatiotemporal feature modeling~\cite{jabri2020space,han2020memory}.

However, there are \lyx{obvious} limitations in these works. Firstly, \lyx{previous works only explore either instance-wise or semantic-wise distribution~\cite{gordon2020watching,qian2020spatiotemporal,li2020prototypical}, lacking a comprehensive perspective over both sides. }
Secondly, \lyx{less effort \js{has been placed} on} low-level features than high-level representations, \lyx{while the former is proven critical for knowledge transfer~\cite{zhao2020makes}}. Third, directly performing temporal augmentations, \emph{e.g.}, shuffle and reverse, at input level \lyx{instead of feature level} could impair feature learning~\cite{bai2020can}.

To address these challenges, we propose a novel framework that explicitly optimizes features from a \emph{\lyx{unified multi-level view}} to achieve more \lyx{general} representations. \lyx{The representations from different levels of deep neural networks show different generalization and abstraction properties. Specifically, \js{it is of common view that} high-level features are more representative towards instances or semantics 
but less 
\js{feasible towards} cross-task transfer. In contrast, low-level features are transfer-friendly but lack structural information over samples, and \js{are particularly} sensitive to temporal statistics.} 

\js{This consideration is particularly meaningful from different perspectives.} \lyx{In a high-level sense,} we optimize the deep representation from two aspects: 1) instance discrimination with conventional InfoNCE loss; 2) semantic structure modeling with a prototypical branch. 
\lyx{In this way, 
the high-level representations \js{can procure} structural relationship among samples} by formulating both instance- and semantic-wise relationships into distribution graphs as depicted in Fig.~\ref{fig:teaser}. \lyx{In a low-level sense, these distribution graphs serve as} reliable \lyx{cues to} aggregate samples that \lyx{share} similar semantics and instance characteristics (\emph{e.g.}, appearance, motion) \js{for better optimization} 
in \lyx{multiple shallower} feature spaces. Through this, \lyx{low-level features are imposed with high-level relation knowledge while keeping good cross-task generalization ability.}

\lyx{Since low-level representation is sensitive to input temporal sequences, we replace the previous data-level temporal augmentation methods with a \js{multi-level} solution to enhance the temporal modeling of the pretrained representation. } 
First, we apply temporal augmentation on multi-level features to construct contrastive pairs that have different motion patterns \lyx{with the objective designed to} distinguish the augmented samples and original ones. Second, a retrieval task \lyx{is proposed} to match the features in short and long \lyx{time spans based on} \js{their} semantic consistency. \lyx{Compared with} previous \lyx{data-level solutions}, our method avoids \js{forcing} the backbone model to adapt to unnatural sequences which corrupts spatiotemporal statistics. Experimental results reveal that our proposed simple temporal modeling \lyx{is more general and suits different network backbones}, while the conventional augmentation technique is somewhat limited to two-pathway networks like SlowFast~\cite{feichtenhofer2019slowfast}.
 
In brief, our contributions can be summarized as:
\begin{itemize}
    \item We propose a multi-level feature optimization framework for unsupervised video representation learning. \js{Both} instance- and semantic-\lyx{wise} knowledge learned from high-level features \js{are leveraged} to form a more reliable \js{self-supervisory} signal, which is employed to optimize low-level feature distributions \js{thereby enhancing} transferability.
    \item 
    We develop a simple but effective temporal modeling module \lyx{with a multi-level augmentation} \js{scheme} for more robust temporal analysis.
    \item Our method achieves state-of-the-art performance on two downstream tasks, action recognition and video retrieval, across two datasets, UCF-101 and HMDB-51. Ablation studies demonstrate the efficacy of multi-level feature optimization as well as the new temporal modeling strategy.
\end{itemize}

\section{Related Work}

\subsection{Contrastive Representation Learning}

Contrastive learning aims to discriminate instances by attracting the positive pairs and repelling the negative pairs~\cite{hadsell2006dimensionality,gutmann2010noise,wu2018unsupervised}. A line of works have adopted this approach for self-supervised representation learning~\cite{hjelm2018learning,oord2018representation,tian2019contrastive,chen2020simple,he2020momentum,ye2019unsupervised}. But there exits one main drawback of the one-hot labels in InfoNCE loss, \emph{i.e.} it only regards the augmentation of the query as positive, and considers all other samples as equally negative. To address this problem, ~\cite{wei2020co2,fang2021seed} employed the similarity distribution in the embedding space to guide contrastive learning in another view. Further, ~\cite{wei2020can,xie2020delving,wang2020unsupervised,han2020unsupervised,li2020prototypical} demonstrated that the semantic-\lyx{wise} relationships between different samples could improve the high-level representation. To better extract the latent semantics in unlabelled data, ~\cite{caron2020unsupervised,asano2020labelling,asano2019self,regatti2020consensus} leveraged Sinkhorn-Knopp algorithm~\cite{cuturi2013sinkhorn} to generate uniformly distributed clusters as pseudo labels for pretraining. However, ~\cite{zhao2020makes} demonstrated \lyx{only utilizing instance discrimination or semantic label is not the optimal solution to} establish transferable representations. Therefore, we propose to jointly consider the instance- and semantic-wise similarity distribution to form a reliable self-supervision signal, which simultaneously maintains the learned instance-wise unique information and filters out hard negatives.

\subsection{Multi-level Feature Analysis}

The features of different layers in the deep neural network tend to possess different attributes, \emph{e.g.}, lower-level features contain more information of object shapes and are more transferable, while higher-level features contain more texture cues and are more specific to certain semantics~\cite{islam2021shape,zou2020revisiting,yosinski2014how,zhao2020makes}. ~\cite{zhao2020makes} demonstrated that it is the low-level and mid-level features that majorly transfer from pretrained networks to downstream tasks. However, most existing works on self-supervised representation learning only focus on high-level features. Though ~\cite{xiong2020loco} attempted to optimize intermediate feature vectors but did not establish relationships between different levels. While in our work, we use joint constraint of instance- and semantic-wise distributions inferred from high-level features to explicitly optimize low-level and mid-level representations, which significantly facilitates pretrained knowledge transfer.

\subsection{Self-supervised Video Representation Learning}

In self-supervised video representation learning, a line of works designed various pretext tasks, \emph{e.g.}, temporal ordering~\cite{misra2016shuffle,xu2019self,yao2020seco}, spatiotemporal puzzles~\cite{kim2019self,wang2020statistic}, colorization~\cite{vondrick2018tracking}, playback speed prediction~\cite{jenni2020video,benaim2020speednet} and temporal cycle-consistency~\cite{wang2019learning,jabri2020space,li2019joint}. Some works proposed to predict future frames from the given sequence to learn feature embeddings~\cite{vondrick2016anticipating,villegas2017decomposing,luo2017unsupervised,behrmann2021unsupervised}. Recently, inspired by the success of contrastive learning in static image, a line of works expanded contrastive learning pipeline to video domain~\cite{gordon2020watching,qian2020spatiotemporal,ma2020towards,wang2020self,liu2021temporal}. Typically, ~\cite{han2019video,han2020memory} employed InfoNCE loss for dense future prediction, ~\cite{kong2020cycle,han2020self} performed instance discrimination across different domains to boost video representation. Though contrastive self-supervised learning contributes to better representation, the temporal information in videos is not well leveraged. ~\cite{bai2020can} revealed that directly applying temporal augmentations on input sequences even impairs the performance since these unnatural sequences could corrupt spatiotemporal statistics. To tackle this problem, ~\cite{wang2020removing,wang2020enhancing} disentangled static appearance and dynamic motion information but required complex training procedures. In contrast, we propose a simple yet effective operation to apply temporal augmentations on extracted multi-level features. In this way, we manage to embed the temporal characteristics to the video backbone without enforcing the network to adapt to unnatural sequences.

\section{Method}

\begin{figure*}
    \centering
    \includegraphics[width=\linewidth]{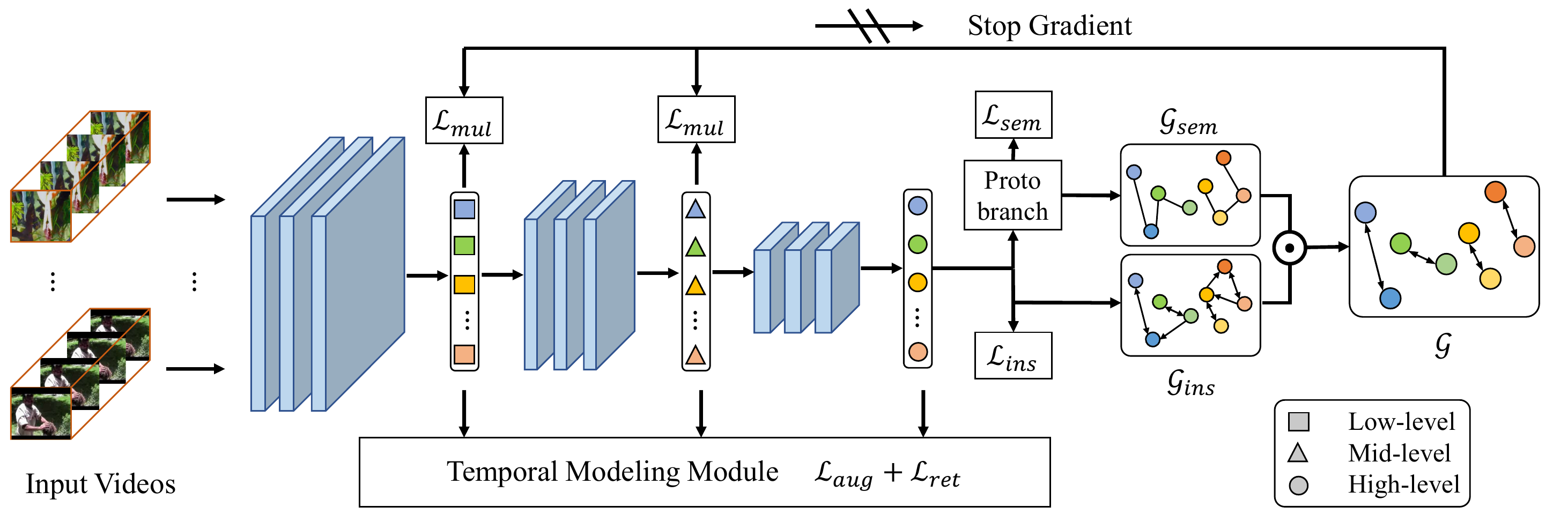}
    \caption{An overview of the multi-level feature optimization framework. We perform instance and semantic discrimination on high-level representations and infer two similarity distribution graphs $\mathcal{G}_{ins}$ and $\mathcal{G}_{sem}$, which are combined into $\mathcal{G}$, a reliable self-supervisory signal to guide low-level and mid-level representation learning. Note that we stop the gradient from back-propagating to the inferred distribution. To \js{exploit} multi-level features of different resolutions, we propose a temporal modeling strategy to enhance motion pattern discrimination.}
    \label{fig:framework}
    \vspace{-1em}
\end{figure*}

In this section, we introduce our proposed multi-level feature optimization framework as shown in Fig.~\ref{fig:framework}. Concretely, we first present our instance and semantic discrimination on high-level representations. Next, we develop the instance- and semantic-\lyx{wise} distribution graph to generate reliable constraint for multi-level feature optimization. Then, we propose a simple temporal modeling approach to improve temporal discrimination at different time scales.

\subsection{Beyond Instance Discrimination}
\label{3.1}
Recent contrastive learning methods based on instance discrimination have shown superior performance on self-supervised representation learning, but the one-hot labels in InfoNCE loss neglect the relationship between different samples. Specifically, as shown in Eq.~\ref{infoNCE},
\begin{align}
    \mathcal{L}_{NCE} = -log\frac{h(\mathbf{q},\mathbf{q}')}{h(\mathbf{q},\mathbf{q}')+\sum_{i=1}^N h(\mathbf{q},\mathbf{k}_i)},
    \label{infoNCE}
\end{align}
where $h(\mathbf{u},\mathbf{v})=\text{exp}(\mathbf{u}^T\mathbf{v}/(\tau||\mathbf{u}||_2||\mathbf{v}||_2))$ with temperature $\tau$, given query $\mathbf{q}$ with its augmentation $\mathbf{q}'$, and a negative key list $\{\mathbf{k}_1,\mathbf{k}_2,...,\mathbf{k}_N\}$, the InfoNCE loss only regards the augmentation of the query as positive and takes all other samples as equally negative. However, considering that existing contrastive self-supervised learning pipelines mostly require large negative pools, there exist some negative samples that \js{may} share similar characteristics, \emph{e.g.}, appearance, motion or category, with the query. Under this circumstance, better instance discrimination would even lead to performance drop in downstream tasks~\cite{tschannen2019mutual}. To this end, besides instance-wise discrimination, we explicitly develop another branch on the projected high-level feature vectors for inter-sample relationship modeling.

Mathematically, we denote the projected high-level feature vector of the $i$-th sample and $a$-th augmentation view as $\mathbf{z}_i^a\in\mathbb{R}^C$, where $C$ is the channel dimension. The instance discrimination learning objective can be formulated as
\setlength{\abovedisplayskip}{4pt}
\setlength{\belowdisplayskip}{5pt}
\begin{align}
    \mathcal{L}_{ins} = -\sum_{i=1}^N\sum_{a=1}^2 log\frac{h(\mathbf{z}_i^1,\mathbf{z}_i^2)}{\sum_{j=1}^N h(\mathbf{z}_i^a,\mathbf{z}_j^*)},
    \label{instance}
\end{align}\vspace{-0.75em}
\begin{align}
    h(\mathbf{z}_i^a,\mathbf{z}_j^*)=\left\{ \begin{array}{lr}
         h(\mathbf{z}_i^1,\mathbf{z}_i^2) & \text{if}\quad i=j \\
         h(\mathbf{z}_i^a,\mathbf{z}_j^1)+h(\mathbf{z}_i^a,\mathbf{z}_j^2) & \text{if}\quad i\neq j
    \end{array}
    \right.
\end{align}
where two augmentation views are adopted, and $N$ is the number of samples within a batch.
For inter-sample relationship modeling, 
\js{we draw motivation from} parametric classification approaches~\cite{cao2020parametric,li2020prototypical} \js{by defining} a learnable matrix $\mathbf{P}\in\mathbb{R}^{C\times K}$ as prototypes to serve as pseudo category centers, where $K$ is the number of prototypes\footnote{$K$ is \js{not required} to be consistent with the number of semantic classes in the training set, it can be set to a comparatively large number as shown in experiments.}. We \js{perform} matrix multiplication between $\mathbf{z}_i^a$ and the prototypes $\mathbf{P}$ followed with softmax regression to produce the 
semantic-wise distribution $\mathbf{p}_i^a\in\mathbb{R}^K$. In the absence of category annotations, it is intuitive to encourage $\mathbf{p}_i$ of different augmentations to be consistent, but it lacks the discrimination between different semantics, which can lead to feature space collapse~\cite{caron2018deep}. Inspired by~\cite{caron2020unsupervised,asano2020labelling,asano2019self} (where clustering is regarded as an optimal transport problem), we employ Sinkhorn-Knopp algorithm~\cite{cuturi2013sinkhorn} to transform a set of distributions $\{\mathbf{p}_1^a,\mathbf{p}_2^a,...,\mathbf{p}_N^a\}$ into soft targets $\{\mathbf{s}_1^a,\mathbf{s}_2^a,...,\mathbf{s}_N^a\}$, \lyx{where $\mathbf{s}^a_i \in \mathbb{R}^{K}$}, is uniformly distributed at category level, \lyx{indicating} that there are around $\frac{N}{K}$ samples per category. In this way, the generated soft targets explicitly discriminate samples of different semantic groups and \lyx{avoids} trivial solutions. Therefore, we optimize the model by minimizing cross-entropy between the soft targets and probability distributions of different augmentations as in Eq~\ref{entropy}:
\begin{align}
    \mathcal{L}_{sem} = -\sum_{i=1}^N\sum_{k=1}^K \mathbf{s}_i^1(k) log\mathbf{p}_i^2(k)+\mathbf{s}_i^2(k) log\mathbf{p}_i^1(k),
    \label{entropy}
\end{align}
where two augmentation views are adopted. Considering that $K$ could be larger than batch size, we design a queue to store the semantic-wise distributions \js{from} previous batches to ensure equal partition into $K$ prototypes, but using only those from the current batch for gradient back-propagation. Different from previous methods~\cite{he2020momentum,caron2020unsupervised,wang2020unsupervised}, we store the inferred distributions in the queue, which \js{would} generally change slower than feature vectors in the training phase. Therefore, our method could work with small batch sizes without requiring a slow-progressing momentum encoder.

Finally, we jointly leverage $\mathcal{L}_{ins}$ and $\mathcal{L}_{sem}$ to form the self-supervisory \js{objective for} high-level representations:
\begin{align}
    \mathcal{L}_{high} = \mathcal{L}_{ins} + \mathcal{L}_{sem}.
    \label{high}
\end{align}
This enables the network to simultaneously discriminate the instances of different characteristics and \js{uncover} potential instance groups that share similar semantics.

\subsection{Graph Constraint for Multi-level Features}

The instance- and semantic-\lyx{wise} constraints lead to effective high-level representations, but it is worth noting that it is the lower-level features that mainly transfer from the pretrained network to downstream tasks~\cite{zhao2020makes}. Therefore, it is crucial to also pay attention to lower-level representations. However, directly applying either instance or semantic discrimination to intermediate layers does not bring improvement~\cite{xiong2020loco,zhao2020makes}, \js{hence} it remains a challenge to impose reasonable \lyx{guidance} on these features. Since we could infer instance- and semantic-\lyx{wise} distributions from high-level features as mentioned in Section~\ref{3.1}, it is intuitive to produce an ideal \js{self-supervisory} signal by taking these two distributions into consideration.

Particularly, we denote the instance-\lyx{wise} similarity distribution as a directed graph $\mathcal{G}_{ins}$, and semantic-\lyx{wise} distribution as an undirected graph $\mathcal{G}_{sem}$. Both two graphs consist of $N$ nodes representing $N$ different samples within a batch, and $N\times N$ edges indicating the relationship between each sample. The detailed formulation of the edges $\mathcal{E}$ is:
\begin{align}
    \mathcal{E}_{ins}(i,j) = \left \{ 
    \begin{array}{lr}
         \mathcal{W}(i,j)\quad & \text{if } \frac{\mathcal{W}(i,j)}{\sum_{j=1}^N\mathcal{W}(i,j)}\geq \eta\\
         0\quad & \text{if } \frac{\mathcal{W}(i,j)}{\sum_{j=1}^N\mathcal{W}(i,j)}<\eta
    \end{array}
    \right.,
    \label{inslevel}
\end{align}
\begin{align}
    \mathcal{E}_{sem}(i,j) = \left \{ 
    \begin{array}{lr}
         1 & \text{if } \text{argmax}(\mathbf{s}_i^*)=\text{argmax}(\mathbf{s}_j^*) \\
         0 & \text{if } \text{argmax}(\mathbf{s}_i^*)\neq\text{argmax}(\mathbf{s}_j^*)
    \end{array}
    \right.,
    \label{semlevel}
\end{align}
\begin{align}
    s.t. \quad\mathcal{W}(i,j) = \left \{ 
    \begin{array}{lr}
         h(\mathbf{z}_i^1,\mathbf{z}_i^2)\quad & \text{if}\quad i=j \\
         \Bar{h}(\mathbf{z}_i^*,\mathbf{z}_j^*)\quad & \text{if}\quad i\neq j
    \end{array}
    \right.,
\end{align}
where two augmentation views are adopted as in Section~\ref{3.1}, and $\eta$ is a threshold hyper-parameter, $\Bar{h}(\mathbf{z}_i^*,\mathbf{z}_j^*)=\frac{1}{4}\sum_{m=1}^2\sum_{n=1}^2h(\mathbf{z}_i^m,\mathbf{z}_j^n)$, $\mathbf{s}_i^*=\mathbf{s}_i^1+\mathbf{s}_i^2$. In this way, $\mathcal{E}_{ins}$ indicates the inferred instance-wise similarity distribution, which respects inter-sample relationship and is more 
realistic data distribution than the one-hot \lyx{encoding}. \lyx{Meanwhile, to filter out} hard negatives that \lyx{share} high similarity in $\mathcal{E}_{ins}$, we employ $\mathcal{E}_{sem}$ to truncate the edges between nodes of different pseudo categories. Under this circumstance, we manage to comprehensively utilize unique instance-wise information and high-level semantics to generate reliable self-supervision for low-level and mid-level features. \textcolor{black}{Mathematically, we jointly leverage $\mathcal{G}_{ins}$ and $\mathcal{G}_{sem}$ to form the combined graph $\mathcal{G}$, whose edge weights $\mathcal{E}$ serve as the final soft targets:} 
\begin{align}
    \mathcal{E}(i,j) = \frac{\mathcal{E}_{ins}(i,j)\mathcal{E}_{sem}(i,j)}{\sum_{k=1}^N\mathcal{E}_{ins}(i,k)\mathcal{E}_{sem}(i,k)}.
    \label{soft}
\end{align}
We then calculate cross entropy between $\mathcal{E}$ and inferred similarity distribution to optimize lower-level features, \textit{i.e.},
\begin{align}
    \mathcal{L}_{mul} =& -\sum_{i=1}^N\sum_{j=1}^N\sum_{a=1}^2 \mathcal{E}(i,j)
    log\frac{h(\mathbf{z_r}_i^a,\mathbf{z_r}_j^*)}{\sum_{j=1}^N h(\mathbf{z_r}_i^a,\mathbf{z_r}_j^*)},
    \label{multi}
\end{align}
where $r$ indicates the feature level (low-level or mid-level) and $\mathbf{z_r}$ is the projected feature vectors of the $r$-th level. With this learning objective, we obtain more robust and representative lower-level features to facilitate knowledge transfer.

\subsection{Temporal Modeling}

Under the proposed multi-level representation optimization framework, it is intuitive to utilize the temporal information at diverse time scales to enhance motion pattern modeling since the features at different layers possess different temporal characteristics. 

Motivated by previous works in video action recognition~\cite{wang2016temporal,zhou2018temporal,piergiovanni2019representation}, \js{achieving} robust temporal modeling \js{entails} two aspects: 1) Semantic discrimination between different motion patterns; 2) Semantic consistency under different temporal views. Therefore, we devise two learning objectives to 
\js{accomplish this.}

First, for motion pattern discrimination, we use general temporal transformations, \emph{e.g.}, temporal shuffle and reverse, to augment samples of various motion patterns. However, since the backbone is learned from scratch, directly applying augmentations on the input data will force the network to adapt to unnatural sequences. 
We develop a simple yet effective operation to perform temporal augmentation on multi-level features $\mathbf{f_r}$, and then leverage a lightweight motion excitation module~\cite{li2020tea} to extract motion enhanced feature representations. Temporal transformations that result in semantically inconsistent motion patterns \lyx{can be regarded} as a negative pair \lyx{of the} original sample \lyx{and the InfoNCE loss is used} to discriminate these augmented pairs, \js{\emph{i.e.}},
\begin{align}
    \mathcal{L}_{aug} &= -\sum_{i=1}^N\sum_{a=1}^2 log\frac{h(\text{ME}(\mathbf{f_r}_i^1),\text{ME}(\mathbf{f_r}_i^2))}{h(\text{ME}(\mathbf{f_r}_i^1),\text{ME}(\mathbf{f_r}_i^2))+\mathbf{neg}_i^a},\\
    s.t.\quad\mathbf{neg}_i^a &= \sum_{k=1}^Ah(\text{ME}(\mathbf{f_r}_i^a),\text{ME}(\text{Aug}_k(\mathbf{f_r}_i^a))),
\end{align}
where $\text{ME}$ is implemented by the Motion Excitation module in~\cite{li2020tea} followed with spatiotemporal average pooling and a two-layer multi-layer perception (MLP), $\text{Aug}_k$ indicates $k$-th temporal augmentation operation. In this way, we embed the ability to discriminate motion patterns into the backbone network. Second, to boost the consistency under different temporal views, we propose to match the feature of a specific timestamp \lyx{from sequences of different lengths}. Concretely, for a short sequence $v_s$ covering timestamp $[t_1,t_2]$ and a long sequence $v_l$ covering $[t_3,t_4]$, where $t_3<t_1<t_2<t_4$, we aim to retrieve the feature at each timestamp of $v_s$ in the feature set of $v_l$. Similarly, we also formulate it as a contrastive learning problem, where the feature of corresponding timestamp in $v_l$ serves as the positive key, while others serve as negatives, \emph{i.e.},
\begin{align}
    \mathcal{L}_{ret} = -\sum_{t_q\in [t_1,t_2]}log\frac{h(v_s(t_q),v_l(t_q))}{\sum_{t_k\in [t_3,t_4]}h(v_s(t_q),v_l(t_k))}.
\end{align}
By leveraging $\mathcal{L}_{aug}$ and $\mathcal{L}_{ret}$, we achieve motion pattern discrimination as well as temporally consistent understanding of different views. Moreover, both learning objectives are implemented on multi-level features with diverse resolutions, leading to more robust temporal modeling.

\section{Experiment}

\subsection{Dataset and Evaluation}

We use three popular video action recognition datasets, Kinetics-400~\cite{carreira2017quo}, UCF-101~\cite{soomro2012ucf101} and  HMDB-51~\cite{kuehne2011hmdb}. For self-supervised pretraining, we use the training set of UCF-101 or Kinetics-400 for fair comparisons. For the downstream tasks, following~\cite{benaim2020speednet,han2020memory,kong2020cycle}, we use split 1 of UCF-101 and HMDB-51 for evaluation.

\begin{table*}[]
\small
\begin{minipage}{0.59\linewidth}
\centering
    \begin{tabular}{c|cccc|cc}
        \hline
        Method & Backbone & Dataset & Res & Freeze & UCF & HMDB \\
        \hline
        CBT~\cite{sun2019learning} & S3D & K600 & 112 & \Checkmark & 54.0 & 29.5 \\
        CCL~\cite{kong2020cycle} & R3D-18 & K400 & 112 & \Checkmark & 52.1 & 27.8 \\
        MemDPC$\dagger$~\cite{han2020memory} & R2D3D-34 & K400 & 224 & \Checkmark & 54.1 & 30.5 \\
        TaCo~\cite{bai2020can} & R3D & K400 & - & \Checkmark & 59.6 & 26.7 \\
        \hline
        Ours & S3D & K400 & 112 & \Checkmark & 61.1 & 31.7 \\
        Ours & R3D-18 & K400 & 112 & \Checkmark & \textbf{63.2} & \textbf{33.4} \\
        \hline
        \hline
        Order~\cite{xu2019self} & R(2+1)D & UCF & 112 & \XSolidBrush & 72.4 & 30.9 \\
        VCP~\cite{luo2020video} & R3D & UCF & 112 & \XSolidBrush & 66.0 & 31.5 \\
        STS~\cite{wang2020statistic} & R3D-18 & UCF & 112 & \XSolidBrush & \textbf{77.8} & 40.7 \\
        PRP~\cite{yao2020video} & R(2+1)D & UCF & 112 & \XSolidBrush & 72.1 & 35.0 \\
        \hline
        Ours & S3D & UCF & 112 & \XSolidBrush & 74.3 & 37.2 \\
        Ours & R3D-18 & UCF & 112 & \XSolidBrush & 76.2 & \textbf{41.1} \\
        \hline
        RotNet~\cite{jing2018self} & R3D-18 & K400 & 112 & \XSolidBrush & 62.9 & 33.7 \\
        CBT~\cite{sun2019learning} & S3D & K600 & 112 & \XSolidBrush & 79.5 & 44.6 \\
        TempTrans~\cite{jenni2020video} & R3D-18 & K400 & 112 & \XSolidBrush & 79.3 & 49.8 \\
        Pace~\cite{wang2020self} & R(2+1)D & K400 & 112 & \XSolidBrush & 77.1 & 36.6 \\
        ST-Puzzle~\cite{kim2019self} & R3D-18 & K400 & 224 & \XSolidBrush & 63.9 & 33.7 \\
        SpeedNet~\cite{benaim2020speednet} & S3D-G & K400 & 224 & \XSolidBrush & \textbf{81.1} & 46.8 \\
        MemDPC~\cite{han2020memory} & R2D3D-34 & K400 & 224 & \XSolidBrush & 78.1 & 41.2 \\
        DSM~\cite{wang2020enhancing} & R3D-34 & K400 & 224 & \XSolidBrush & 78.2 & \textbf{52.8} \\
        \hline
        Ours & S3D & K400 & 112 & \XSolidBrush & 76.5 & 42.3 \\
        Ours & R3D-18 & K400 & 112 & \XSolidBrush & 79.1 & 47.6 \\
        \hline
    \end{tabular}
    \caption{Comparison results for action recognition task. We show of settings of the backbone used, pretraining dataset, resolution for fair comparison. Freeze (tick) indicates linear probe, while no freeze (cross) indicates the finetune mode. $\dagger$ means using two-stream networks, $i.e.$, RGB and optical flow.}
    \label{tab:action}
\end{minipage}
\hspace{0.05\linewidth}
\begin{minipage}{0.33\linewidth}  
\centering
    \begin{tabular}{c|c|cc}
        \hline
        Method & Multi-level & UCF & HMDB \\
        \hline
        w/o & \XSolidBrush & 58.1 & 28.8 \\
        One-hot & \Checkmark & 57.8 & 28.5 \\
        Instance & \Checkmark & 59.4 & 30.1 \\
        Semantic & \Checkmark & 54.5 & 26.3 \\
        Combined & \XSolidBrush & 60.4 & 32.3 \\
        Combined & \Checkmark & \textbf{63.2} & \textbf{33.4} \\
        \hline
    \end{tabular}
    \caption{Ablation study on multi-level feature optimization. Results are based on R3D-18.\\\\}
    \label{tab:graph}
    \begin{tabular}{c|c|cc}
        \hline
        Method & Backbone & UCF & HMDB \\
        \hline
        w/o TM & SlowFast & 57.8 & 30.3 \\
        w/ CTM & SlowFast & \textbf{65.4} & \textbf{34.8} \\
        w/ NTM & SlowFast & 63.9 & 34.1 \\
        \hline
        w/o TM & R3D-18 & 55.9 & 28.1 \\
        w/ CTM & R3D-18 & 43.2 & 21.1 \\
        w/ NTM & R3D-18 & \textbf{63.2} & \textbf{33.4} \\
        \hline
        w/o TM & S3D & 53.8 & 27.2 \\
        w/ CTM & S3D & 41.1 & 19.8 \\
        w/ NTM & S3D & \textbf{61.1} & \textbf{31.7} \\
        \hline
    \end{tabular}
    \caption{Ablation study on temporal modeling. Results are shown on three backbones: one two-pathway network SlowFast, two single-pathway networks (R3D-18, S3D). TM: temporal modeling, CTM: convention temporal modeling approach, NTM: our temporal modeling strategy.}
    \label{tab:temporal}
\end{minipage}
\end{table*}

\subsection{Implementation Details}

\noindent\textbf{Self-supervised Pretraining.} We use R3D-18~\cite{hara2017learning} or S3D~\cite{xie2018rethinking} as the backbone network. For temporal augmentation, we use temporal shuffle and reverse as two typical transformations. For the definition of contrastive pairs, we regard clips from the same video as positive pairs, and those of different videos as negative. Specifically, we randomly sample 32 RGB frames within a video, and uniformly split them into two 16-frame clips with resolution $112\times 112$ to form positive pairs. For the proposed timestamp retrieval, we regard 16-frame clips as short sequences and the 32-frame clips as long sequences. For multi-level feature optimization, we formulate it as a two-stage procedure. In the first few epochs, we only use Eq.~\ref{high} to optimize high-level features until they could generate reliable soft targets in Eq.~\ref{soft}. Then, we jointly use Eq.~\ref{high} and Eq.~\ref{multi} for multi-level feature learning. The specific definition of multi-level features is listed in the Supplementary Material. We use batch size of 256, and set default number of prototypes to 1000 with queue length 1024. In total, we train for 100 epochs on Kinetics-400, and 300 epochs on UCF-101 using ADAM with an initial learning rate of $10^{-3}$ and weight decay of $10^{-5}$. The learning rate is decayed by 10 at 70 epochs for Kinetics-400, and 200 epochs for UCF-101.

\noindent\textbf{Action Recognition.} For action recognition, we initialize the backbone with pretrained model parameters except the last fully-connected layer. There are two settings for this task: 1) Finetune the whole network in a fully supervised manner (denoted as \textit{finetune}); 2) Only train the linear classifier (denoted as \textit{linear probe}). For evaluation, following~\cite{xu2019self,wang2020self}, we uniformly sample 10 clips for each video, then center crop and resize them to $112\times 112$. The final prediction of each video is the average softmax probabilities of each clip. Performance is measured by Top-1 accuracy.

\noindent\textbf{Video Retrieval.} For video retrieval, we directly use the pretrained model as a feature extractor without finetuning. Following~\cite{xu2019self,luo2020video}, we select videos in test set as query, and aim to retrieve k-nearest neighbors in training set. We employ the cosine similarity in feature space to measure the similarity, and use Top-k recall R@k for evaluation.


\begin{table*}[]
\small
    \centering
    \begin{tabular}{c|cc|cccc|cccc}
        \hline
        \multirow{2}{*}{Method} & \multirow{2}{*}{Backbone} & \multirow{2}{*}{Dataset} & \multicolumn{4}{c|}{UCF-101} & \multicolumn{4}{c}{HMDB-51} \\
        \cline{4-11}
         & & & R@1 & R@5 & R@10 & R@20 & R@1 & R@5 & R@10 & R@20 \\
        \hline
        SpeedNet~\cite{benaim2020speednet} & S3D-G & Kinetics-400 & 13.0 & 28.1 & 37.5 & 49.5 & - & - & - & - \\
        VCP~\cite{luo2020video} & R3D & UCF-101 & 18.6 & 33.6 & 42.5 & 53.3 & 7.6 & 24.4 & 36.3 & 53.6 \\
        Pace~\cite{wang2020self} & R3D-18 & UCF-101 & 23.8 & 38.1 & 46.4 & 56.6 & 9.6 & 26.9 & 41.1 & 56.1 \\
        MemDPC~\cite{han2020memory} & R2D3D-34 & UCF-101 & 20.2 & 40.4 & 52.4 & 64.7 & 7.7 & 25.7 & 40.6 & 57.7 \\
        PRP~\cite{yao2020video} & R3D & UCF-101 & 22.8 & 38.5 & 46.7 & 55.2 & 8.2 & 25.8 & 38.5 & 53.3 \\
        DSM~\cite{wang2020enhancing} & I3D & UCF-101 & 17.4 & 35.2 & 45.3 & 57.8 & 7.6 & 23.3 & 36.5 & 52.5 \\
        STS~\cite{wang2020statistic} & R3D-18 & UCF-101 & 38.3 & \textbf{59.9} & 68.9 & 77.2 & 18.0 & 37.2 & 50.7 & \textbf{64.8} \\
        \hline
        Ours & R3D-18 & UCF-101 & \textbf{39.6} & 57.6 & \textbf{69.2} & \textbf{78.0} & \textbf{18.8} & \textbf{39.2} & \textbf{51.0} & 63.7 \\
        \hline
        Ours & R3D-18 & Kinetics-400 & 41.5 & 60.6 & 71.2 & 80.1 & 20.7 & 40.8 & 55.2 & 68.3 \\
        \hline
    \end{tabular}
    \caption{Comparison results for video retrieval task. We report R@k (k=1,5,10,20) on UCF-101 and HMDB-51 datasets.}
    \label{tab:retrieval}
    \vspace{-1.0em}
\end{table*}

\subsection{Evaluation on Downstream Tasks}

\noindent\textbf{Action Recognition.} In this subsection, we compare our method with recent state-of-the-art self-supervised video representation learning approaches on video action recognition. In Table~\ref{tab:action}, we report the Top-1 accuracy of two settings, \emph{i.e.}, linear probe and finetune. We exclude models with much deeper backbones or multi-modal data from our comparison. Note that~\cite{he2019rethinking} reports that with various training settings in finetune \js{mode}, even training from scratch could reach performances comparable \js{to that} using pretrained models. Therefore, the linear probe can \js{more consistently} compare the learned representations.%

Under the linear probe setting, our method obtains the best results on both datasets. Specifically, our method with S3D and R3D-18 backbones outperform contrastive learning based approaches, CBT~\cite{sun2019learning} and CCL~\cite{kong2020cycle}, respectively, by a large margin. Even when \lyx{compared} with MemDPC~\cite{han2020memory} which leverages two stream information (RGB and flow), with larger resolution, our method still shows significant \js{advantages}. Additionally, our method also outperforms TaCo~\cite{bai2020can}, an approach that carefully designs various temporal pretext tasks, \lyx{demonstrating} our model's \js{ability to represent the temporal aspect generally.}

Under the end-to-end finetune setting, for models pretrained on UCF-101, our method can outperform approaches that used simple temporal order or playback rate as their pretext task, and is comparable to STS~\cite{wang2020statistic} which designed a complex learning scheme to \js{characterize} appearance and motion statistics. This demonstrates that our method \js{is capable of} robust spatiotemporal modeling. For models pretrained on Kinetics dataset, our method is comparable to recent state-of-the-art approaches even with smaller resolution; \js{this is evident of} the \lyx{generalization} of learned representations. Note that due to limited computational resources, we only report results with resolution 112 and training epochs 100. According to ~\cite{qian2020spatiotemporal,wang2020statistic}, further improvement is expected when using resolution 224 and more epochs for self-supervised pretraining.


\noindent\textbf{Video Retrieval.} Besides the video action recognition task, we also report the video retrieval performance. Table~\ref{tab:retrieval} shows the quantitative results on UCF-101 and HMDB-51. Our method pretrained on UCF-101 is superior to other approaches over two datasets. Our method is significantly better than those using temporal cues to design pretext tasks. Though DSM~\cite{wang2020enhancing} and STS~\cite{wang2020statistic} designed 
elaborate operations to build static appearance and dynamic motion statistics, \js{our higher performance indicates good transferability of knowledge to the downstream task, hence} showing the efficacy of our multi-level feature optimization. \lyx{Further improvement can be observed when Kinetics-400 is utilized}.


\subsection{Ablation Study}
\label{4.5}
Here, we present ablation studies on key modules in the framework as well as some crucial experiment settings. We report the results on action recognition under linear probe setting to evaluate the learned video representations.

\noindent\textbf{Multi-level Optimization.} In this work, we use the graph constraint in Eq.~\ref{soft} to guide lower-level feature learning. We compare it with using different constraints for lower-level features: one-hot labels in Eq.~\ref{infoNCE}, only instance-wise distribution in Eq.~\ref{inslevel}, only semantic-wise distribution in Eq.~\ref{semlevel}, and no constraint \emph{i.e.}, only $\mathcal{L}_{high}$ for high-level features. Besides, we also compare with using combined graph constraint on high-level features as an extra loss term. Results on R3D-18 for these different settings are shown in Table~\ref{tab:graph}. We regard the method that only optimizes high-level representations without any constraints as baseline. When using one-hot labels (in InfoNCE loss) as self-supervision for lower-level features, the poorer-than-baseline performance can be explained by 
gradient competition~\cite{romero2014fitnets,xiong2020loco}. Using $\mathcal{E}_{ins}$ improves the results, while $\mathcal{E}_{sem}$ only appears to badly corrupt the learned representations. This is because $\mathcal{E}_{ins}$ is a soft probability distribution, the learning process is similar to distilling knowledge from high-level representations. On the contrary, $\mathcal{E}_{sem}$ is a hard 0-1 distribution and enforces invariance between samples of the same inferred category. Our method employing \lyx{both constraints} shows significant improvement as combining both distributions \js{can yield} reliable self-supervision. We also observe that introducing graph constraint on high-level representations does bring improvement but still less effective than our full pipeline. This shows that the multi-level feature optimization produces more transferable representations.

\noindent\textbf{Temporal Modeling.} We compare our temporal modeling approach with the conventional temporal augmentation technique, which shuffles or reverses input video clips to construct contrastive pairs, on three backbones (R3D-18, S3D, and two-pathway network SlowFast). Table~\ref{tab:temporal} shows that the conventional approach improves the action recognition performance of SlowFast network, but not the performance of R3D and S3D. This is because the model is trained from scratch, it needs to learn robust spatiotemporal statistics from the input data. For R3D and S3D that use a single 3D convolution pathway to learn 3D features, the temporally shuffled or reversed clips \js{may exhibit} different spatiotemporal statistics \lyx{from} \js{what is deemed as natural}, thus corrupting the learned representations. SlowFast's explicit disentanglement of 
static appearance and dynamic motions 
allows temporally augmented clips of different motion patterns to \js{thrive well}. 
In contrast, our proposed temporal modeling method performs augmentation and discrimination in the \lyx{multi-level} feature space. \lyx{The augmentation part,} which is particularly analogous to 
the projection head of~\cite{chen2020simple}, \lyx{is only utilized during training, \js{hence} the parameters do not affect backbone inference, avoiding} unnatural sequences. To sum up, our simple temporal modeling operation is effective for both single-pathway and two-pathway backbones, while the conventional approach might be limited to two-pathway networks.
\begin{figure}
    \centering
    \includegraphics[width=\linewidth]{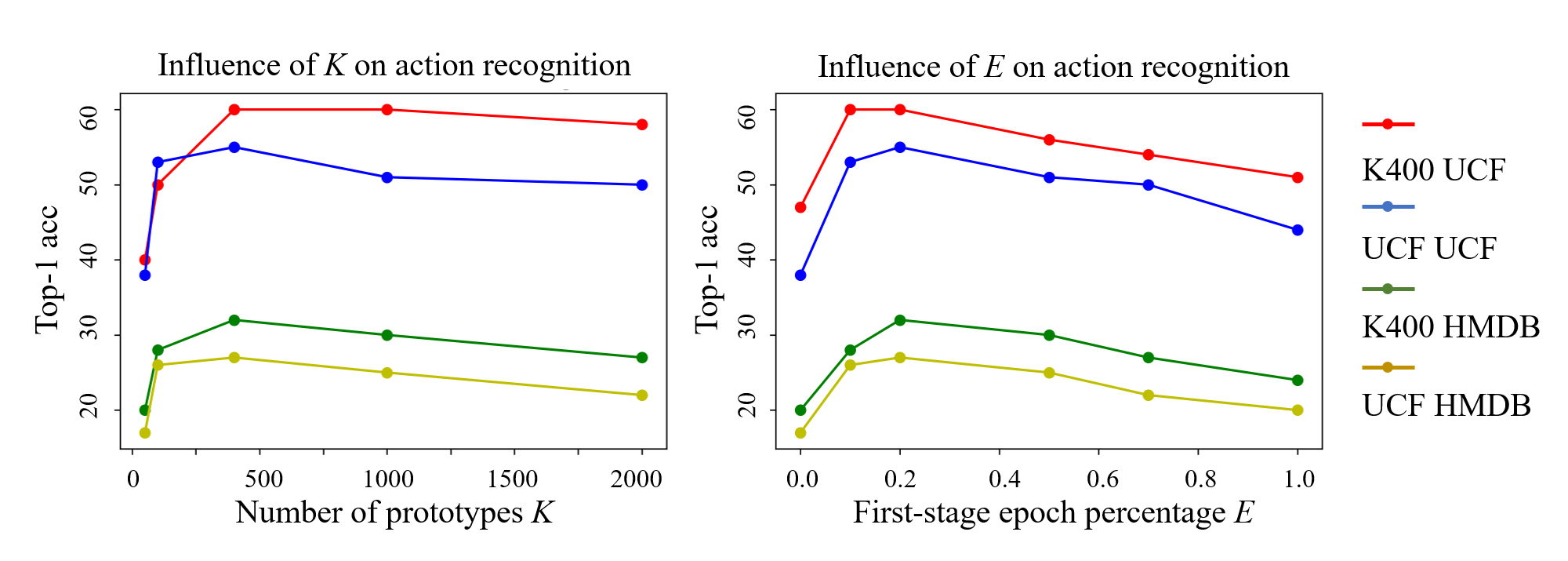}
    \caption{Ablation study on two hyper-parameters: $K$ and $E$. The legend presents pretrained dataset and action recognition evaluation dataset, \emph{e.g.}, the red line denotes pretraining on Kinetics-400 and evaluation on UCF-101.}
    \label{fig:ablation}
    \vspace{-1em}
\end{figure}

\noindent\textbf{Number of Prototypes.} We explore the influence of the hyper-parameter $K$, the number of prototypes, on action recognition. We show the results pretrained on UCF-101 and Kinetics-400 with \js{a range of $K$ values} in Fig.~\ref{fig:ablation}. It demonstrates that it is not necessary to set $K$ equal to the number of categories of a specific dataset, \emph{e.g.}, 101 on UCF, 400 on K400. Instead, by setting $K$ to a relatively larger number, the Top-1 accuracy is still comparatively high. It is worth noting that if $K$ is too small (especially smaller than the number of categories), the performance drops significantly. Because when $K$ is too small, the learned semantic-\lyx{wise} discrimination is too coarse-grained and fails to filter out the hard negatives when formulating the graph constraint. In summary, it is not difficult to set a reasonable value for $K$ for pretraining. A comparatively large number is enough, and we set $K=1000$ as default.

\noindent\textbf{Two-stage Training Split.} We formulate our multi-level feature optimization as a two-stage process. At the first stage, we only optimize high-level features to obtain good initialization of instance- and semantic-\lyx{wise} distribution. At the second stage, we jointly optimize all multi-level features. Here, we explore the influence of the \js{stage} split by training epochs. Fig.~\ref{fig:ablation} compares different first stage training portion as percentage $E$ of epochs, and we make several observations. First, when $E=0$, $i.e.$, no first stage training, the supervision for lower-level features (in Eq.~\ref{soft}) is randomly initialized and this \js{derails} feature learning. Second, when $E$ is large, optimization on lower-level features are not sufficient, hence \js{weaker} transferability of learned representations affects retrieval performance. Third, the performance is best when $E$ is within the range of $[10\%,20\%]$, a stable range for both UCF-101 and Kinetics-400.

\begin{figure}
    \centering
    \includegraphics[width=\linewidth]{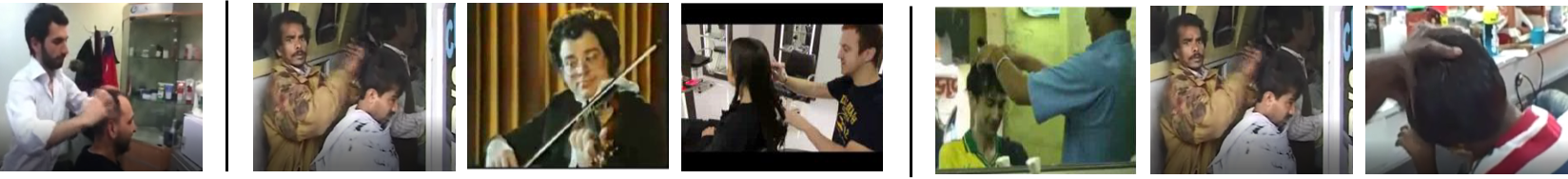}
    \includegraphics[width=\linewidth]{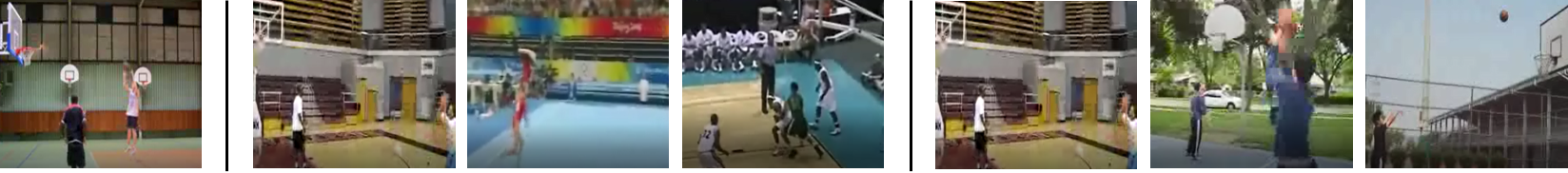}
    \caption{Retrieval of Top-3 similar samples in two distributions. Left: Query sample; Middle: Top-3 samples of instance-wise distribution, Right: Top-3 samples of semantic-wise distribution. 
    }
    \label{fig:dis}
    \vspace{-1em}
\end{figure}
\begin{figure}
    \centering
    \subfigure[Results of our temporal modeling approach.]{
    \includegraphics[width=0.23\linewidth]{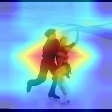}
    \includegraphics[width=0.23\linewidth]{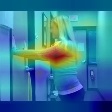}
    \includegraphics[width=0.23\linewidth]{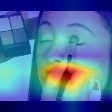}
    \includegraphics[width=0.23\linewidth]{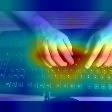}
    }
    \subfigure[Results of conventional temporal modeling approach.]{
    \includegraphics[width=0.23\linewidth]{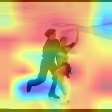}
    \includegraphics[width=0.23\linewidth]{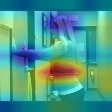}
    \includegraphics[width=0.23\linewidth]{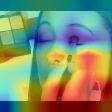}
    \includegraphics[width=0.23\linewidth]{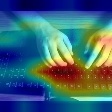}
    }
    \caption{CAM visualization of important motion areas. We use the heatmap to reveal how much temporal cues are contained in each spatial grid. \js{Our approach learns these areas well.}}
    \label{fig:temporal}
    \vspace{-0.9em}
\end{figure}

\subsection{Qualitative Analysis}

Based on the inferred instance- and semantic-\lyx{wise} similarity distributions, we \lyx{list} the Top-3 most similar samples from each distribution based on the example query in Fig.~\ref{fig:dis}. The results show that instance-\lyx{wise} similarity distribution provides samples with similar appearance or motion characteristics, while semantic-\lyx{wise} distribution provides samples of the same \lyx{semantic} category. \lyx{The intersection} of these two distributions leads to sample pairs that share both spatiotemporal characteristics and semantics. This demonstrates that the combined graph constraint could serve as a reliable \js{self-supervisory signal that} maintains unique instance-wise information and is able to discriminate different semantics.

To evaluate the temporal modeling performance, we use the pretrained backbone as a feature extractor, and train a linear classifier to discriminate temporally \lyx{augmented} features from the original. We provide the CAM~\cite{zhou2016learning} visualization in Fig.~\ref{fig:temporal} to show how much temporal cues are contained in each region. It is clear that our temporal modeling strategy contributes to more accurate and discriminative motion areas, while conventional temporal augmentation \js{is not able to perceive important motion cues well}. For example, our method precisely focuses on the moving hands in the typing scene, but the conventional approach regards the keyboard as the motion key.

\section{Conclusion}

In this work, we propose a multi-level feature optimization framework for unsupervised video representation learning. We perform instance- and semantic-\lyx{wise} discrimination on high-level features, thereby employing 
reliable \js{self-supervisory cues} to optimize lower-level representations \js{for improved} \lyx{generalization}. \lyx{Meanwhile}, we also leverage multi-level features of various \lyx{temporal spans} for robust temporal modeling. Extensive experiments demonstrate that our learned representations achieve superior performance on \lyx{a series of downstream tasks.} 

\noindent\textbf{Acknowledgement} The paper is supported in part by the following grants: National Key Research and Development Program of China Grant (No.2018AAA0100400), National Natural Science Foundation of China (No. 61971277), and our corporate sponsors.

{\small
\bibliographystyle{ieee_fullname}
\bibliography{submission}
}

\end{document}